\documentclass[letterpaper, 10 pt, conference]{ieeeconf}
\usepackage[utf8]{inputenc}
\IEEEoverridecommandlockouts
\pdfoutput=1

\usepackage[british]{babel}

\usepackage[T1]{fontenc}

\usepackage{graphicx}
\graphicspath{{figures/}}

\usepackage{amsmath,amssymb,amsfonts}

\usepackage[hyphens]{url}
\usepackage[colorlinks=true, allcolors=blue]{hyperref}
\usepackage{balance}
\usepackage{arydshln}

\usepackage[noadjust]{cite}

\title{\LARGE \bf
Extended Abstract: 4D Radar Gaussian Modeling and Scan Matching with RCS}

\author{
Fernando Amodeo, \textit{Graduate Student Member, IEEE}$^{1}$, Luis Merino, \textit{Member, IEEE}$^{1}$ and Fernando Caballero$^{1}$%
\thanks{$^{1}$Service Robotics Laboratory -- Universidad Pablo de Olavide (Seville), Spain
{\tt\small \{famozur, lmercab, fcaballero\}@upo.es}}
}

\begin{document}

\maketitle
\thispagestyle{empty}
\pagestyle{empty}

\section{Introduction}


4D millimeter-wave (mmWave) radars are increasingly used in robotics, as they offer robustness against adverse environmental conditions.
Besides the usual XYZ position, they provide Doppler velocity measurements as well as Radar Cross Section (RCS) information for every point. While Doppler is widely used to filter out dynamic points \cite{ekfrio}, RCS is often overlooked and not usually used in modeling and scan matching processes. Building on previous 3D Gaussian modeling and scan matching work \cite{gaussians2026}, we propose incorporating the physical behavior of RCS in the model, in order to further enrich the summarized information about the scene, and improve the scan matching process.


\section{Related work}

Point cloud registration is a well known component of odometry, localization and SLAM systems. There exist well known algorithms such as NDT, GICP and VGICP \cite{ndt,gicp,vgicp} that provide reasonable results for many different kinds of point clouds. However, they are all purely based on geometry (point positions), without taking into account additional non-geometric information.

There are 4D radar works incorporating RCS \cite{lessismore,raislam,vgcrio}, however they are all focused on matching individual points between a reference point cloud and an incoming cloud; and only in local timeframes, as they make the assumption that the RCS values remain unchanged due to the small change in view direction. In our case we are interested in enhancing an existing 3D Gaussian representation for 4D radar data \cite{gaussians2026} by incorporating a more general physical model of the RCS behavior of the points that fall within the Gaussian, explicitly taking into account the view direction to calculate a predicted RCS value that can guide scan matching.

\section{Methodology}

RCS is a distance-independent quantity that provides information about the reflectivity of the surface to the radar waves. It depends on surface properties such as the material, texture or shape; and as such, it is similar to view-dependent color information (lighting behavior). Inspired by 3D Gaussian Splatting (3DGS) \cite{3dgs}, we adopt Spherical Harmonics (SH) to model the behavior of RCS. SH are a generalization of the Fourier series that works in two periodic dimensions (the surface of a sphere), and allows for representing an arbitrary continuous function.

\subsection{Gaussian RCS modeling}

Using an already built 3D Gaussian model of a scene \cite{gaussians2026}, we augment the information of each individual Gaussian by calculating a set of SH coefficients that approximate the RCS values returned by the radar. In particular, we select the points corresponding to a given Gaussian, then calculate and normalize the incidence vector for each point (defined as the difference between the point and the center of the Gaussian in the local radar frame). We ensure the vector always faces the radar by forcing the X component to be negative (that is, flipping the vector if the X is positive). We then define a new vector consisting of the outputs from all SH functions up to a given degree, using the incidence vector as input. The predicted RCS value is the dot product between the RCS coefficients and the SH function outputs, with the coefficients being directly derived through linear least squares.

The raw RCS value from the radar is not used as-is, as it is noisy and often in wildly different scales. We first perform a statistical analysis of the RCS values of all points within a given Gaussian, finding the median value. We then find outliers by defining an interval below/above the median based on the minimum absolute difference between the median and the minimum/maximum RCS value. Points outside this interval are considered outliers and thus discarded, and the raw RCS values are transformed into the $[-1, 1]$ interval by subtracting the median and dividing by the discovered scale. This is done in order to ensure that residuals are all in the same order of magnitude later on during scan matching.

\subsection{RCS-enhanced scan matching}

\begin{figure*}[t]
\centering
\includegraphics[width=0.86\textwidth]{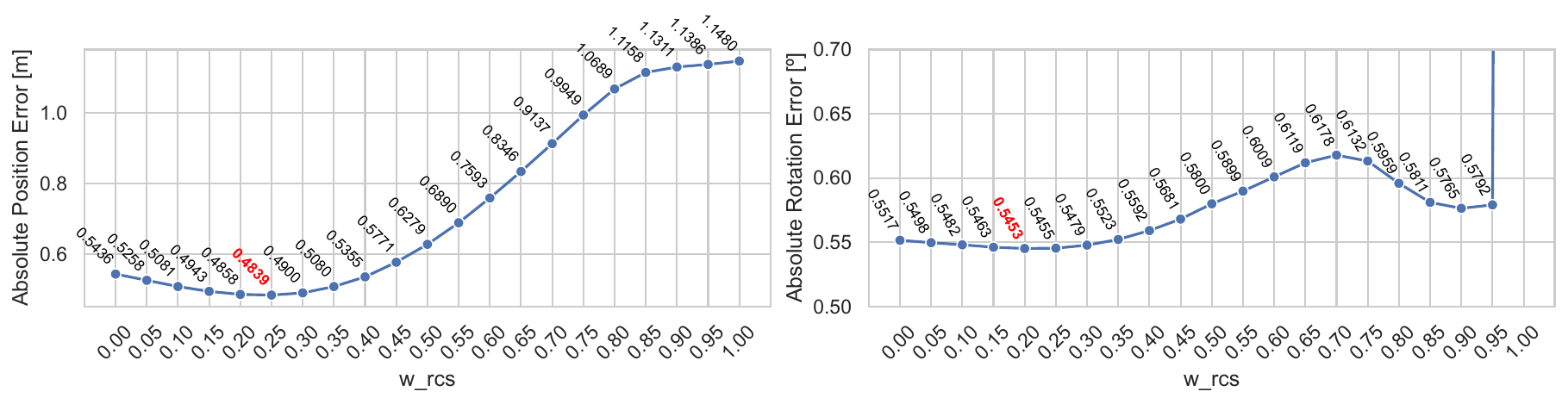}
\caption{Preliminary localization results on Snail-Radar's \texttt{20231213/data1} sequence. Best results in red and bold.}
\label{pufo}
\end{figure*}

Scan matching is typically performed by a second order optimization algorithm such as Gauss-Newton. The fundamental equation to solve is:
\begin{equation}
\mathbf{H}\delta\mathrm{T}=-\nabla f,
\end{equation}
where $f$ is the cost function, $\nabla f$ is the gradient of $f$, $\mathbf{H}$ is the Hessian approximation of $f$, and $\delta \mathrm{T}$ is the $\mathfrak{se}(3)$ change vector to apply to the transformation towards the solution. The cost function of conventional scan matching is purely geometric, as it is based on the position of points or distributions with respect to the modeled points or distributions; and thus we call it $f_\mathrm{geo}$. We propose another cost function $f_\mathrm{rcs}$ that measures the mean square error of the RCS approximation produced by the model with respect to the real RCS value measured by the radar. The approximation function is the same defined by the previous section. We also process the square residuals using the Cauchy loss function in order to robustify the optimization process.

An important thing to note is that while the entire pose is optimized for the geometrical part (both translation and rotation), we propose only optimizing the rotation part for the RCS part. This is because the rows and columns of the Hessian matrix corresponding to translation contain vanishingly small values, which causes the determinant to be very close to zero, making the matrix effectively singular. Therefore, we propose forcing those rows/columns to be the identity matrix, with the corresponding gradient forced to be zero.

Finally, we define a combined cost function,
\begin{equation}
f = \frac{1 - w_\mathrm{rcs}}{N_\mathrm{geo}} f_\mathrm{geo} + \frac{w_\mathrm{rcs}}{N_\mathrm{rcs}} f_\mathrm{rcs};
\end{equation}
incorporating both geometric and RCS components, and scaling each by the number of points that contributed to each part (note that the RCS part excludes outliers). There is a hyperparameter $w_\mathrm{rcs}$ controlling the importance of RCS.

\section{Preliminary results}

In order to test the effects of incorporating RCS data in modeling and scan matching, we construct a Gaussian-RCS model of the \texttt{20231213/data1} sequence from the Snail-Radar dataset \cite{snailradar} based on radar scans produced by the ARS548 radar, which directly provides RCS information. We perform the well known Doppler RANSAC-LSQ filtering for dynamic points \cite{ekfrio} and merge all scans from the sequence using the ground truth poses in order to build a single point cloud that is passed to our Gaussian-RCS modeling algorithm. Afterwards, we iterate through every scan, and perform scan matching using the ground truth plus additive translational/rotational noise against the generated model. We perform a hyperparameter search for $w_\mathrm{rcs}$, ranging from 0 (RCS-less baseline) to 1 (only using RCS without geometric information). We use SH functions of degrees 0 through 3, for a total of 16 SH coefficients per Gaussian. The random translation/rotation noise added to the initial pose estimates is distributed uniformly between 0 and 2\,m; and between 0\textdegree\,and 5\textdegree\,respectively.

Fig.~\ref{pufo} shows the results. We can observe that a value of $w_\mathrm{rcs} = 0.25$ produces the best overall absolute position errors, and it is also very close to the best absolute rotation errors. An interesting thing to note is that despite RCS only being used to optimize the rotation, it seems to have a regularizing effect in the position. Finally, it seems the RCS cost function is not suitable to be used alone without the geometric component ($w_\mathrm{rcs} = 1$), which is expected as pure rotation optimization alone cannot provide results when there is a known translational error.

\section{Conclusions and future work}

Our results provide support to the idea that radar RCS information can be incorporated into Gaussian models, and that its use can bring improvements in a localization task.

Regarding future work, we intend to continue developing the RCS system to further enhance its efficacy, as well as integrating it into an odometry system that incrementally builds a map of the environment.

\balance

\bibliographystyle{IEEEtran}
\bibliography{IEEEabrv,allthecites}

\begin{thebibliography}{10}
\providecommand{\url}[1]{#1}
\csname url@rmstyle\endcsname
\providecommand{\newblock}{\relax}
\providecommand{\bibinfo}[2]{#2}
\providecommand\BIBentrySTDinterwordspacing{\spaceskip=0pt\relax}
\providecommand\BIBentryALTinterwordstretchfactor{4}
\providecommand\BIBentryALTinterwordspacing{\spaceskip=\fontdimen2\font plus
\BIBentryALTinterwordstretchfactor\fontdimen3\font minus \fontdimen4\font\relax}
\providecommand\BIBforeignlanguage[2]{{%
\expandafter\ifx\csname l@#1\endcsname\relax
\typeout{** WARNING: IEEEtran.bst: No hyphenation pattern has been}%
\typeout{** loaded for the language `#1'. Using the pattern for}%
\typeout{** the default language instead.}%
\else
\language=\csname l@#1\endcsname
\fi
#2}}

\bibitem{ekfrio}
C.~Doer and G.~F. Trommer, ``{An EKF Based Approach to Radar Inertial Odometry},'' in \emph{{2020 IEEE International Conference on Multisensor Fusion and Integration for Intelligent Systems (MFI)}}, 2020, pp. 152--159.

\bibitem{gaussians2026}
F.~Amodeo, L.~Merino, and F.~Caballero, ``{4D Radar-Inertial Odometry Based on Gaussian Modeling and Multi-Hypothesis Scan Matching},'' \emph{IEEE Robotics and Automation Letters}, vol.~11, no.~5, pp. 5773--5780, 2026.

\bibitem{ndt}
P.~Biber and W.~Strasser, ``{The normal distributions transform: a new approach to laser scan matching},'' in \emph{{Proceedings 2003 IEEE/RSJ International Conference on Intelligent Robots and Systems (IROS 2003) (Cat. No.03CH37453)}}, vol.~3, 2003, pp. 2743--2748 vol.3.

\bibitem{gicp}
A.~Segal, D.~Hähnel, and S.~Thrun, ``{Generalized-ICP},'' in \emph{{Robotics: Science and Systems}}.\hskip 1em plus 0.5em minus 0.4em\relax The MIT Press, 2009.

\bibitem{vgicp}
K.~Koide, M.~Yokozuka, S.~Oishi, and A.~Banno, ``{Voxelized GICP for Fast and Accurate 3D Point Cloud Registration},'' in \emph{2021 IEEE International Conference on Robotics and Automation (ICRA)}, 2021, pp. 11\,054--11\,059.

\bibitem{lessismore}
Q.~Huang, Y.~Liang, Z.~Qiao, S.~Shen, and H.~Yin, ``{Less is More: Physical-Enhanced Radar-Inertial Odometry},'' in \emph{2024 IEEE International Conference on Robotics and Automation (ICRA)}, 2024, pp. 15\,966--15\,972.

\bibitem{raislam}
D.~C. Herraez, M.~Zeller, D.~Wang, J.~Behley, M.~Heidingsfeld, and C.~Stachniss, ``{RaI-SLAM: Radar-Inertial SLAM for Autonomous Vehicles},'' \emph{IEEE Robotics and Automation Letters}, vol.~10, no.~6, pp. 5257--5264, 2025.

\bibitem{vgcrio}
J.~Xiang, X.~He, Z.~Chen, L.~Zhang, X.~Luo, and J.~Mao, ``{VGC-RIO: A Tightly Integrated Radar-Inertial Odometry With Spatial Weighted Doppler Velocity and Local Geometric Constrained RCS Histograms},'' \emph{IEEE Robotics and Automation Letters}, vol.~10, no.~11, pp. 11\,642--11\,649, 2025.

\bibitem{3dgs}
\BIBentryALTinterwordspacing
B.~Kerbl, G.~Kopanas, T.~Leimk{\"u}hler, and G.~Drettakis, ``{3D Gaussian Splatting for Real-Time Radiance Field Rendering},'' \emph{{ACM Transactions on Graphics}}, vol.~42, no.~4, July 2023. [Online]. Available: \url{https://repo-sam.inria.fr/fungraph/3d-gaussian-splatting/}
\BIBentrySTDinterwordspacing

\bibitem{snailradar}
\BIBentryALTinterwordspacing
J.~Huai, B.~Wang, Y.~Zhuang, Y.~Chen, Q.~Li, and Y.~Han, ``{SNAIL radar: A large-scale diverse benchmark for evaluating 4D-radar-based SLAM},'' \emph{The International Journal of Robotics Research}, vol.~0, no.~0, p. 02783649251329048, 0. [Online]. Available: \url{https://journals.sagepub.com/doi/abs/10.1177/02783649251329048}
\BIBentrySTDinterwordspacing

\end{thebibliography}

\end{document}